\documentclass[acmtog,authorversion,nonacm]{acmart}
\acmSubmissionID{344}

\usepackage{booktabs} 
\usepackage{caption}
\usepackage{subcaption}
\usepackage{tikz}
\usepackage{xcolor}

\usepackage[ruled]{algorithm2e} 

\SetAlFnt{\small}
\SetAlCapFnt{\small}
\SetAlCapNameFnt{\small}
\SetAlCapHSkip{0pt}

\acmJournal{TOG}

\begin{document}
\title{DistGrid: Scalable Scene Reconstruction with Distributed Multi-resolution Hash Grid}

\author{Sidun Liu}
\orcid{0000-0001-7715-4698}
\affiliation{%
 \institution{National University of Defence Technology}
 \city{Changsha}
 \country{China}}
\email{liusidun@nudt.edu.cn}
\author{Peng Qiao}
\orcid{0000-0001-6752-7892}
\affiliation{%
 \institution{National University of Defence Technology}
 \city{Changsha}
 \country{China}
}
\email{pengqiao@nudt.edu.cn}
\author{Zongxin Ye}
\affiliation{%
 \institution{National University of Defence Technology}
 \city{Changsha}
 \country{China}
}
\author{Wenyu Li}
\affiliation{%
 \institution{National University of Defence Technology}
 \city{Changsha}
 \country{China}
}
\author{Yong Dou}
\orcid{0000-0002-1256-8934}
\affiliation{%
 \institution{National University of Defence Technology}
 \city{Changsha}
 \country{China}
}
\email{yongdou@nudt.edu.cn}

\renewcommand\shortauthors{Liu, S. et al}

\begin{abstract}
    Neural Radiance Field~(NeRF) achieves extremely high quality in object-scaled and indoor scene reconstruction. However, there exist some challenges when reconstructing large-scale scenes. MLP-based NeRFs suffer from limited network capacity, while volume-based NeRFs are heavily memory-consuming when the scene resolution increases. Recent approaches propose to geographically partition the scene and learn each sub-region using an individual NeRF. Such partitioning strategies help volume-based NeRF exceed the single GPU memory limit and scale to larger scenes. However, this approach requires multiple background NeRF to handle out-of-partition rays, which leads to redundancy of learning. Inspired by the fact that the background of current partition is the foreground of adjacent partition, we propose a scalable scene reconstruction method based on joint Multi-resolution Hash Grids, named DistGrid. In this method, the scene is divided into multiple closely-paved yet non-overlapped Axis-Aligned Bounding Boxes, and a novel segmented volume rendering method is proposed to handle cross-boundary rays, thereby eliminating the need for background NeRFs. The experiments demonstrate that our method outperforms existing methods on all evaluated large-scale scenes, and provides visually plausible scene reconstruction. The scalability of our method on reconstruction quality is further evaluated qualitatively and quantitatively.
\end{abstract}

%
%


%
%


\begin{teaserfigure}
    \centering
    \begin{subfigure}[b]{0.431\textwidth}
        \centering
        \includegraphics[width=\textwidth]{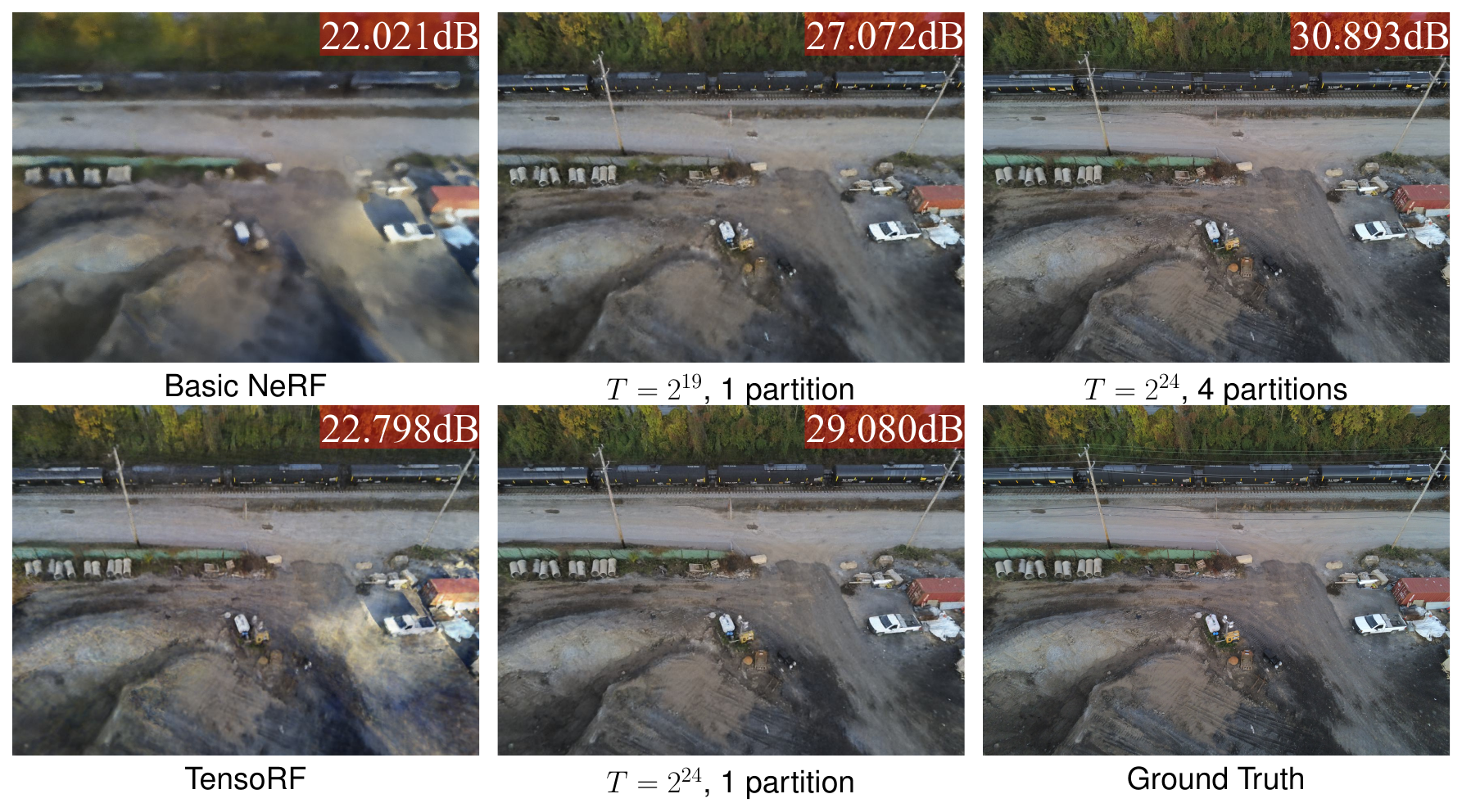}
        \caption{Visual comparison.}
        \label{fig:visual_scalability}
    \end{subfigure} 
    \tikz{\draw[-,blue, densely dashed, thick](0,-2.4) -- (0,2.4);}
    \begin{subfigure}[b]{0.289\textwidth}
        \centering
        \includegraphics[width=\textwidth]{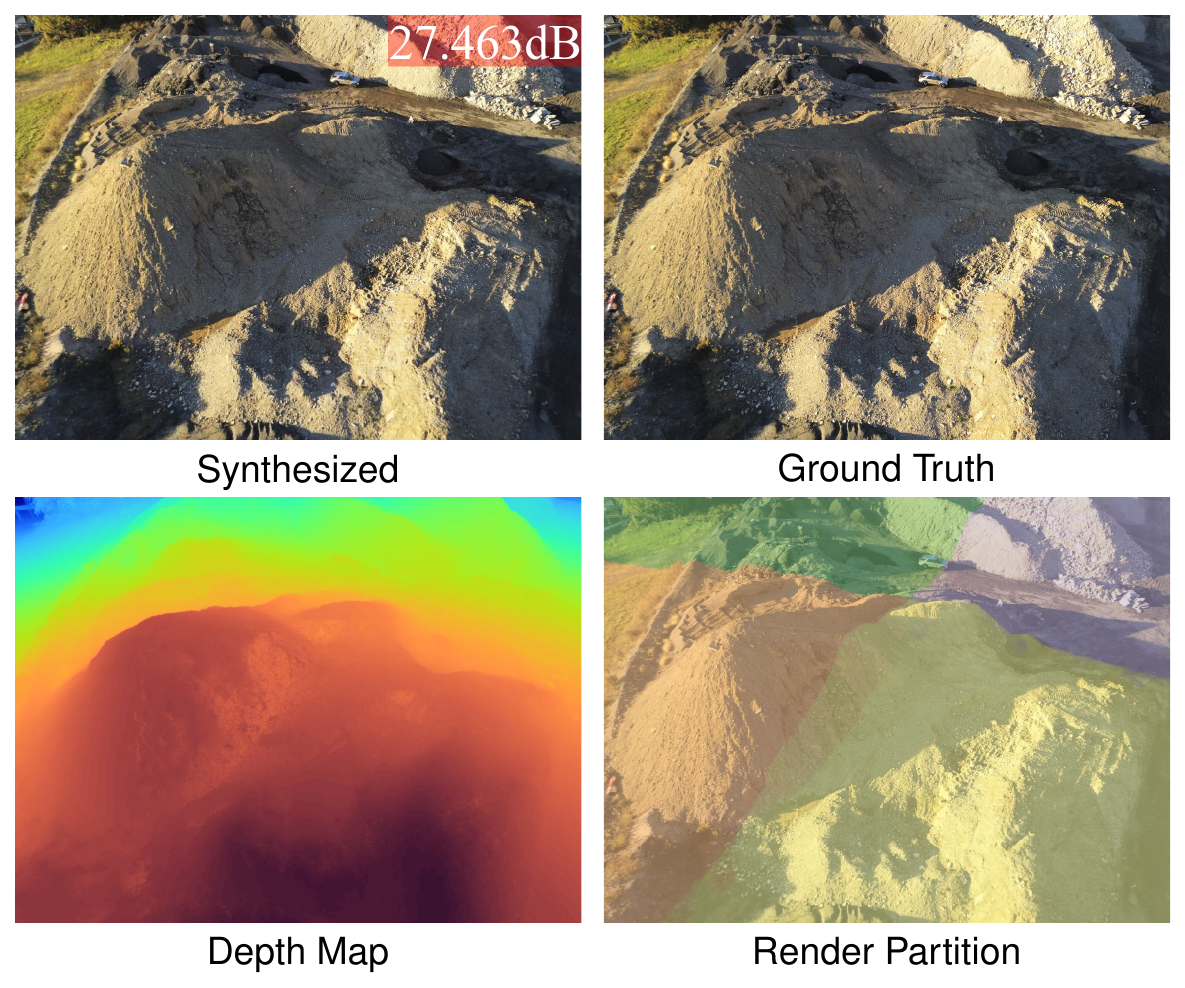}
        \caption{Novel view on the boundary.}
        \label{fig:boundary_render}
    \end{subfigure}
    \tikz{\draw[-,blue, densely dashed, thick](0,-2.4) -- (0,2.4);}
    \begin{subfigure}[b]{0.25\textwidth}
        \centering
        \includegraphics[width=\textwidth]{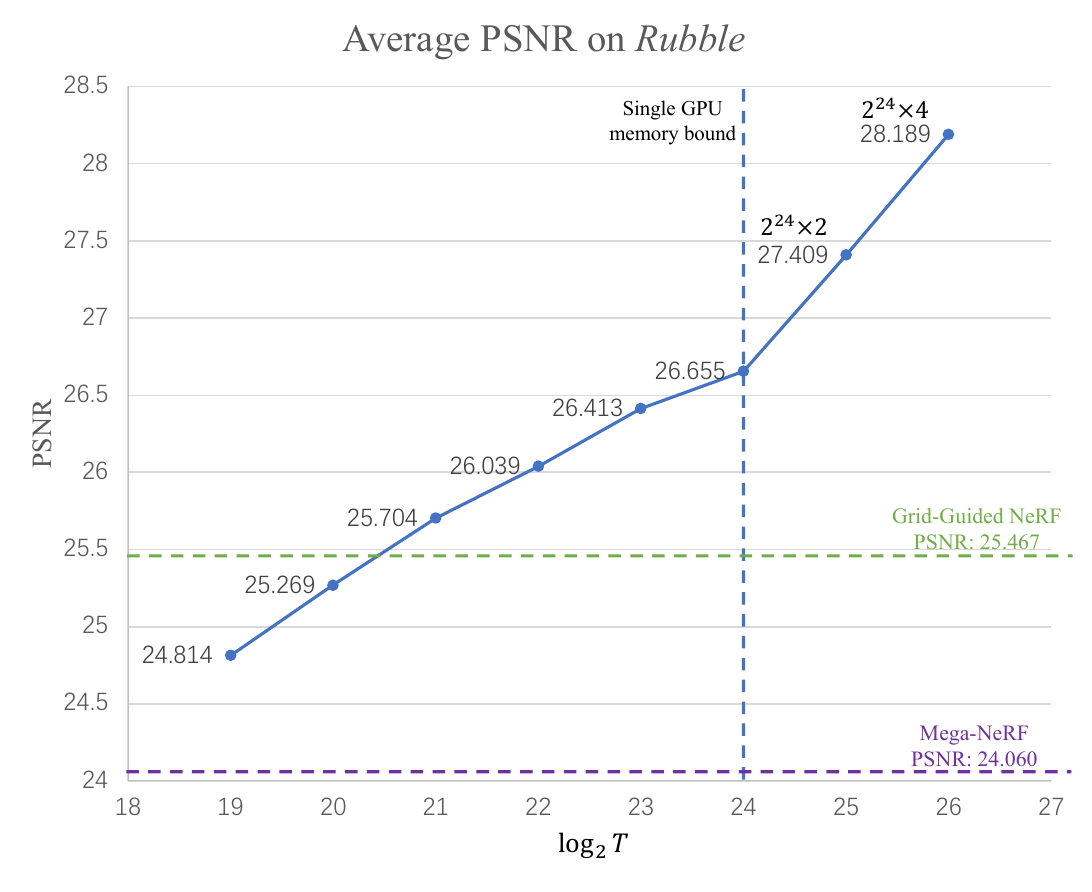}
        \caption{The PSNRs to hash table length.}
        \label{fig:rubble_plot}
    \end{subfigure}
      \caption{We scale up the Multi-resolution Hash Grid to enable large-scale scene reconstruction in a distributed manner. (a) The visual quality of synthesized novel view images from test set, with varying hash table length, compared to basic NeRF~\cite{mildenhall2021nerf} and TensoRF~\cite{chen2022tensorf} models. Synthesis quality increases with larger hash tables. However, the memory of a single GPU limits its further growth, which is bounded under $T=2^{24}$. Therefore, we partition the scene to enable distributed reconstruction. (b) The scene is partitioned into closely-paved yet non-overlapped bounding boxes. Dispite the partitioning, there are no artifacts on the boundaries due to the proposed Segmented Volume Rendering method. The different colors in \textit{Render Partition} indicate that they are rendered by different sub-models. (c) The curve of average PSNR on dataset \textit{Rubble}, with varying hash table length and scene partitions~(e.g. $2^{24}\times 2$ means the scene is split into 2 parts and $T=2^{24}$ is applied for each partition). The performance improves as the length of hash table increases. Even though the hash tables are distributed across different GPUs, the growth trend is maintained. DistGrid improves the evaluation PSNR of \textit{Rubble} dataset up to \textbf{2.7 dB} with native Multi-resolution Hash Grid models, compared to presently state-of-the-art method~\cite{xu2023grid}.}
      \label{fig:teaser}
\end{teaserfigure}

\maketitle

\section{Introduction}

Neural rendering technology has made significant progress since the proposal of Neural Radiance Field (NeRF)~\cite{mildenhall2021nerf}, which is designed to address the \textit{novel view synthesis} task~\cite{chen1993view,shum2000review}. Based on volume rendering~\cite{max1995optical}, it represents a scene implicitly with a multi-layer perceptron (MLP) called a neural field. This neural field accepts 3D coordinates and viewing directions as input and predicts their corresponding density and color. NeRF produces more than impressive visual quality on the \textit{novel view synthesis} task. Its learned neural field can also be used for scene reconstructions~\cite{schonberger2016structure}. Subsequent works are derived from NeRF to improve efficiency and quality~\cite{lin2021barf,barron2021mip,muller2022instant,chen2022tensorf}.

However, the implicit representation limits NeRF to object-scale or indoor scene reconstruction, and is not applicable for larger-scale scenes, such as landscapes or cityscapes, due to limited network capacity. 
While Mega-NeRF~\cite{turki2022mega} aims to address these problems by partitioning the scene into different regions and training individual NeRF++~\cite{zhang2020nerf++} sub-models for each sub-region, it still requires more than one day of training on eight high-end GPUs.

Volume-based hybrid representation~\cite{fridovich2022plenoxels,sun2022improved,muller2022instant,chan2022efficient,chen2022tensorf} has been identified as a superior solution compared to implicit representation-based methods. These methods consume more memory in exchange for training efficiency, making them scalable for reconstructing larger scenes. GP-NeRF~\cite{zhang2023efficient} and Grid-Guided NeRF~\cite{xu2023grid} leaverage Multi-resolution Hash Grid (MHG)~\cite{muller2022instant} and Multi-Plane Image (MPI)~\cite{chan2022efficient,chen2022tensorf} respectively to enhance the model capacity and training efficiency for large-scale scene reconstruction.

\begin{figure*}[t]
    \centering
    \includegraphics[width=0.92\textwidth]{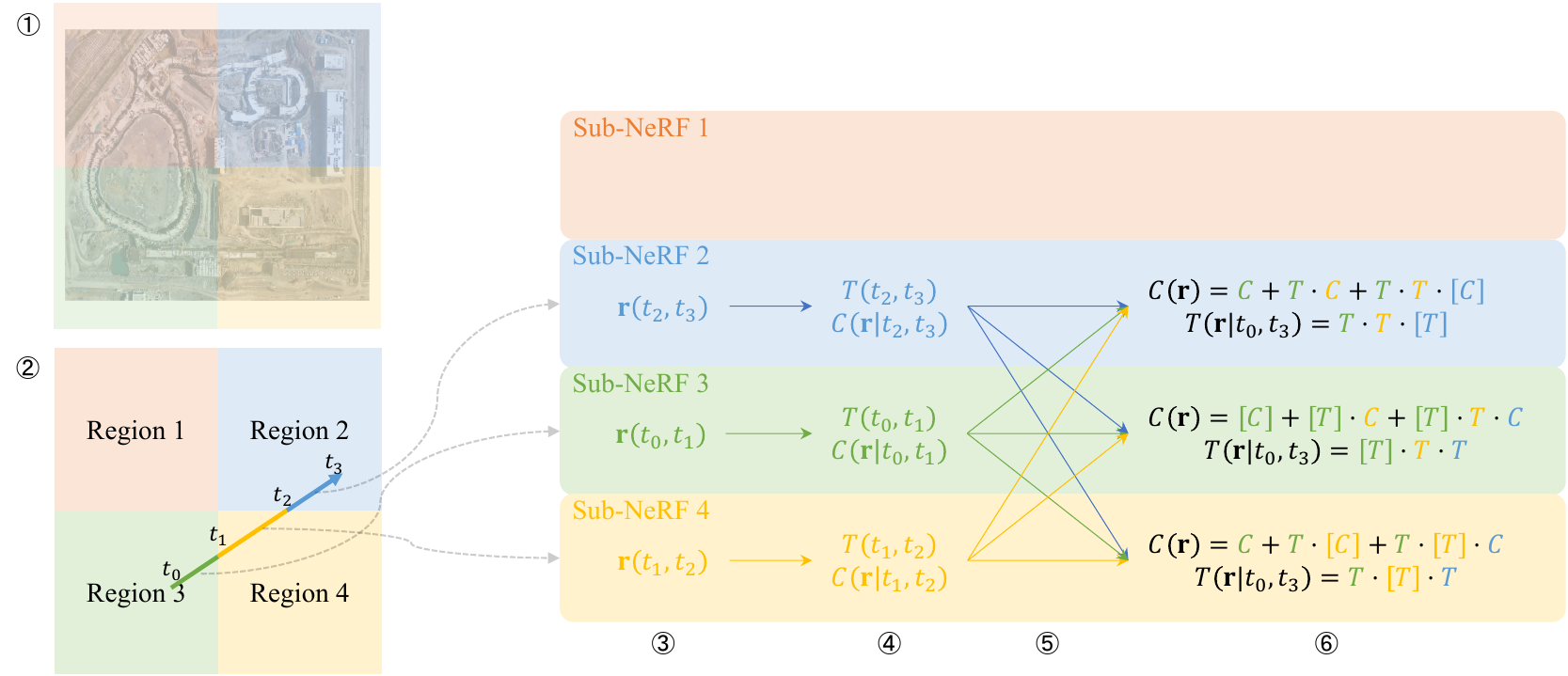}
    \caption{This figure illustrates the segmented volume rendering process, taking 4-partition as an example. Cross-region rays are rendered locally and merged globally. \textcircled{1} The scene is partitioned into 4 spatial regions, with each region reconstructed by a sub-NeRF model. \textcircled{2} Input rays are first tested using Ray-Intersection with AABBs to obtain its start, end, and cross-boundary moments ($t_0\sim t_3$). \textcircled{3} Rays are then segmented and distributed to corresponding sub-NeRF models. \textcircled{4} Local volume rendering is performed to obtain local transmittance and color. \textcircled{5} The local results are scattered in the group of sub-NeRF models that the ray intersects. \textcircled{6} Finally, based on the segmented volume rendering equations (\ref{eq:color_fwd}) and (\ref{eq:transmittance_fwd}), the gathered results are merged. Note that only the locally computed transmittance and color (marked with $[\cdot]$) require gradients in back-propagation.}
    \label{fig:seg_vren}
\end{figure*}

These methods trade off memory usage for training efficiency, thereby their scalability is limited by single GPU memory size. Although it is feasible to split the scene into sub-regions and train them separately using the partitioning strategy from Mega-NeRF~\cite{turki2022mega}, this strategy is not compatible with MHG due to two issues. Firstly, wrapping the scene with several spheres (or ellipsoids) causes region overlaps. Secondly, two NeRF models are trained to represent foreground and background respectively, but in practice, the background of the current region is the foreground of the adjacent region. Both issues result in redundancy, which can negatively impact training efficiency.

We believe that the fundamental reason for these issues is that the sub-NeRFs are trained in isolation, without communication between sub-regions. To overcome this challenge and develop a scalable scene reconstruction method, we propose DistGrid based on joint MHGs. In our approach, the scene is partitioned into closely-paved yet non-overlapped Axis-Aligned Bounding Boxes (AABBs), and individual sub-NeRFs are trained for each of the sub-regions. To facilitate this partitioning strategy, the cubic bounding box used in Instant NGP~\cite{muller2022instant} is extended to the deformable bounding box.

To handle cross-region rays, we propose a segmented volume rendering method in which the rays are rendered locally and then merged globally to generate the final rendering results. This method allows the adjacent regions to serve as the background of each other, avoiding redundant training of background NeRF models.
We have also ovserved that oblique photography datasets often cover a large area beyond the central region. To address this issue, a two-level cascade structure is adopted in each sub-NeRF. Here, a fine-level MHG is applied to reconstruct the central area, while a coarse-level MHG covers the outer area. An example of partitioning is provided in Fig.~\ref{fig:partition}.
These sub-NeRFs can be deployed on different GPUs, creating a distributed system. Since the inter-GPU bandwidth demand is relatively small~(refer to Sec.~\ref{sec:SegmentedVren}), the training efficiency in a distributed context is not significantly affected.

Our experiments on large-scale scene demonstrate the scalability of our proposed method. The reconstruction quality improves with the increase of the model size. Moreover, with the help of our joint training scheme, consistent reconstruction quality can be achieved on the boundary under the non-overlap setting.

\section{Related Work}

\subsection{Large-Scale Scene Reconstruction}

Large-scale scenes refer to outdoor environments or scenes that cover a vast area, such as landscapes, cityscapes, or even entire cities. These scenes typically require a much larger amount of data to be processed and a more complex representation to capture their details accurately.

Reconstructing large-scale scenes is a long-standing problem. Traditional methods~\cite{agarwal2011building,pollefeys2008detailed} adopt the Structure from Motion (SfM)~\cite{schonberger2016structure} pipeline to solve it. Given a set of scene pictures, their camera poses are estimated with the pipeline of feature extraction~\cite{ping2013review}, feature matching~\cite{sarlin2020superglue}, and epipolar geometry solving~\cite{zhang1998determining}. Then, the dense multi-view stereo~\cite{ishikawa1999mapping} and triangular meshing~\cite{botsch2010polygon} are applied to generate the 3D mesh model of the scene.

Neural Radiance Field (NeRF)~\cite{mildenhall2021nerf,wang2021neus} exploit deep learning techniques for reconstruction. A series of works have extended NeRF to handle large-scale scenes. 
NeRFusion~\cite{zhang2022nerfusion} progressively construct and fuse local volumes to build the global large indoor volumes. 
BungeeNeRF~\cite{xiangli2022bungeenerf} adopts a residual architecture and progressively enlarges the network for multi-scale city scene reconstruction. 
Urban Radiance Field~\cite{rematas2022urban} leverages LiDAR data and RGB-based sky segmentation method for better reconstruction of street-view environments. 
Mixture-of-Experts technique~\cite{masoudnia2014mixture} is applied in LoE-NeRF~\cite{hao2022implicit} and Switch-NeRF~\cite{mi2023switchnerf} to enlarge the capacity of the network to improve large-scale reconstructions.

The idea of divide-and-conquer has also been employed in recent works to decompose large-scale scenes. Block-NeRF~\cite{tancik2022block} divides the street view into several blocks and represents them by individual NeRFs. Mega-NeRF~\cite{turki2022mega} divides the scene based on the camera positions to render a large-scale scene captured by a drone. SUDS~\cite{turki2023suds} decomposes the scene into individually trained models, each with its own dynamic NeRF.
In the above methods, each sub-model is trained in isolation without considering the connection between partitions.

Beyond MLP-based NeRF, several works have used volumetric representations to tackle large-scale scenes. GP-NeRF~\cite{zhang2023efficient} adopts MHG and a ground feature plane to achieve faster scene reconstruction. Grid-Guided NeRF~\cite{xu2023grid} uses a multi-resolution ground feature plane to coarsely capture the scene, and complement it with another NeRF to achieve higher quality.
However, their performance is limited by the volume resolution.

\subsection{Volumetric Scene Representation}

In previous works, the most common methods used for volumetric scene representation are voxel grid~\cite{choy20163d,seitz1999photorealistic} and multi-plane images (MPIs)~\cite{zhou2018stereo,srinivasan2019pushing}. Voxel grid is flexible to represent arbitrary geometry but can be memory limited at high resolution. Octree~\cite{Knoll2006ASO} structure is designed for memory reduction.

Entering the deep learning era, volume-based implicit representations are used for modeling occupancy~\cite{chen2019learning,mescheder2019occupancy} and signed distance field~\cite{park2019deepsdf}. After NeRF~\cite{mildenhall2021nerf} shows up, many works replace MLP-based neural field with differentiable volumetric encoders. 
Plenoxels~\cite{fridovich2022plenoxels} uses a sparse voxel grid without neural networks to explicitly represent the scene. 
DVGO~\cite{chan2022efficient} attaches a shallow MLP after a dense voxel grid for better reconstruction details. 
To handle the memory problem of high-resolution voxel grid, Instant NGP~\cite{muller2022instant} proposes a multi-resolution voxel grid and maps the voxels to a hash table, without considering the hash collision. Instant NGP achieves high volume resolution and fast training speed, but its performance is bounded by the hash table length. Based on the MPI method, EG3D~\cite{chan2022efficient} proposes to use a tri-planar representation for human face 3D structure. TensoRF~\cite{chen2022tensorf} seeks low-rank representation of voxel grid and decomposes it into several 2D feature matrices and 1D feature vectors.

\section{Preliminary}

\subsection{Volume Rendering}

Volume rendering~\cite{max1995optical} is a method to render 2D images from 3D volume in a differentiable way, by compositing the colors of 3D points on a ray into the ray's color. Given a ray $\mathbf{r}=\mathbf{o}+t\mathbf{d}$ emitted from camera center $\mathbf{o}$ through a given pixel on the image plane with direction $\mathbf{d}$, the color $C(\mathbf{r})$ of the ray $\mathbf{r}$ is:

\begin{equation}
    \label{eq:standard_vren}
    \begin{aligned}
        C(\mathbf{r}) = \int_{t_n}^{t_f}T(t)\sigma(\mathbf{r}(t))c(\textbf{r}(t), \mathbf{d})dt, \\\text{where } T(t)=\exp\left(-\int_{t_n}^t(\sigma(\mathbf{r}(s))ds\right)
    \end{aligned}
\end{equation}
$t_n$ and $t_f$ are pre-defined near and far bounds of the scene. NeRF approximates the integral with discrete sampling. The network parameters are optimized by minimizing the square error of composited color $C(\mathbf{r})$ and actual pixel color $C_{gt}(\mathbf{r})$:

\begin{equation}
    \label{eq:standard_loss}
    \mathcal{L} = \sum_i||C_{gt}(\mathbf{r}_i)-C(\mathbf{r}_i)||_2^2
\end{equation}
where $i$ travels over all pixels of the image collection.

\subsection{Multi-resolution Hash Grid}

NeRF~\cite{mildenhall2021nerf} uses an MLP to parameterize the network, leading to low training efficiency and limited frequency. Therefore, hybrid representations are proposed to alleviate these problems. Multi-resolution Hash Grid~(MHG)~\cite{muller2022instant} is an efficient hybrid representation of a scene. It uses multiple levels of cubic voxel grids with different resolutions. In each level, the vertices of the grid are mapped to the entries of a linear hash table with a fixed length, without considering hash collision. When querying the embedding of 3D points in the cube, trilinear interpolation and hash mapping are performed at each level and these embeddings are concatenated and fused with a shallow MLP. The parameter count is bounded by $L\cdot T\cdot F$, where $L$ is the number of levels, $T$ is the hash table length, and $F$ is the embedding size in each level. Empirically, a common GPU~(e.g. RTX 3090 24G) supports the configuration of $L=16,F=2,T=2^{24}$ at most due to memory limitation.

Besides the hash grid, an occupancy grid is maintained to record the density of each voxel in the cube, helping skip the empty areas during ray marching, which greatly improves the ray sampling efficiency.

\section{Approach}

In our proposed DistGrid, we first extend the cubic MHG to make it deformable (Sec.~\ref{sec:DeformableMHG}). Then we partition the scene into closely-paved yet non-overlapped AABBs and put sub-NeRFs into each of them (Sec.~\ref{sec:Partition}). After that, the proposed segmented volume rendering method is used to handle the cross-region rays (Sec.~\ref{sec:SegmentedVren}). Finally, these sub-NeRFs are trained jointly (Sec.~\ref{sec:training}).

\subsection{Deformable Multi-resolution Hash Grid}
\label{sec:DeformableMHG}

\begin{figure}
    \centering
    \includegraphics[width=0.4\textwidth]{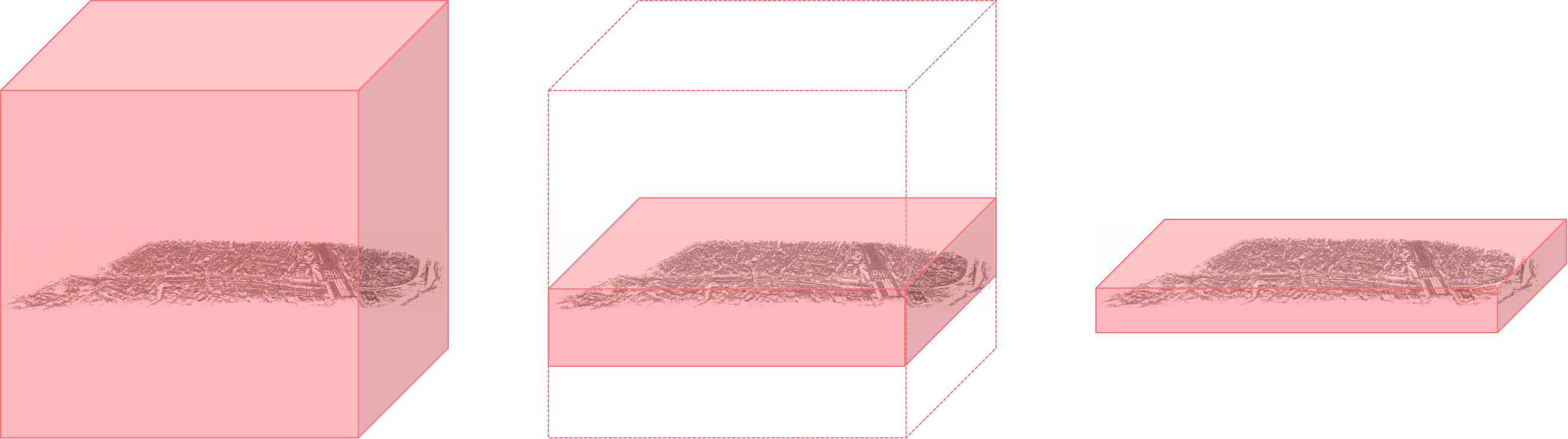}
    \caption{Deformable Multi-resolution Hash Grid. Instant NGP uses a cubic bounding box (\textbf{left}) to wrap the scene, leading to an unnecessary sampling of high-altitude and underground areas. Logical implementation where altitude range is limited (\textbf{center}) still causes additional memory usage. Therefore, the cubic bounding box is extended so that it has an arbitrary aspect ratio (\textbf{right}).}
    \label{fig:deformable}
\end{figure}

The MHG was first proposed in Instant NGP~\cite{muller2022instant} for object-scale or indoor scene reconstruction, where cubic bounding boxes are suitable. However, in the case of large-scale scenes, a non-cubic bounding box may be more appropriate. 

Using a cubic bounding box to wrap the scene (Fig.~\ref{fig:deformable}, left) results in an unnecessary sampling of high-altitude and underground areas during ray marching and can negatively affect efficiency. Referring to Mega-NeRF, a non-cubic bounding box could be implemented by limiting the altitude range of ray sampling (Fig~\ref{fig:deformable}, center). However, this approach requires additional memory usage. In Instant NGP, a one-to-one map is used instead of a hash map if the voxel number is less than the hash table length at the current level in order to save memory. The voxel number in this solution reaches the hash table length earlier in the levels, resulting in additional memory cost.

Therefore, a deformable bounding box (Fig.~\ref{fig:deformable}, right) becomes necessary. This provides us with greater flexibility when performing scene partitioning while also optimizing memory usage. Given the resolution $N_l$ at level $l$, and the aspect ratio of bounding box $(a, b, c)$, the grid shape is represented by $\left(\lceil\frac{a}{s}N_l\rceil,\lceil\frac{b}{s}N_l\rceil,\lceil\frac{c}{s}N_l\rceil\right)$, where $\lceil\cdot\rceil$ is round up operator and $s=\max(a,b,c)$ is the scale factor. The occupancy grid is extended to a deformable version as well. 

\subsection{Scene Partitioning}
\label{sec:Partition}

\subsubsection{Region Partitioning}

Based on the scalability experiments conducted on the \textit{Rubble} dataset (Fig.~\ref{fig:rubble_plot}), we observed a positive correlation between reconstruction quality and hash table length for MHG. However, GPU memory size limits the achievable hash table length. To address this limitation, we propose partitioning the scene and distributing sub-regions across different GPUs. This allows for the entire hash table to be divided and distributed into separate GPU memories.

As the scene exhibits small altitude variance relative to latitude and longitude, DistGrid decomposes it into a horizontal 2D grid similar to Mega-NeRF~\cite{turki2022mega}. However, the decomposed regions in DistGrid are represented as AABBs instead of ellipsoids. These AABBs are closely-paved yet non-overlapped and share the same altitude range while having different latitude and longitude ranges. A ray can be tested through Ray-Intersection with AABBs~\cite{woo1990fast} at runtime to determine which region it belongs to and then distributed to the corresponding device using inter-GPU communication. This process can be efficiently executed without the need for ray preprocessing. If a ray crosses multiple regions, it should be distributed to all of these regions and trained jointly, which will be discussed in Sec.~\ref{sec:SegmentedVren}.

\subsubsection{Coarse-Fine Partitioning}

Given the ground plane altitude and camera poses, these cameras' fields of view (FOV) are projected onto the ground plane to obtain the reconstruction area. We wrap this area with an axis-aligned box. Oblique photography makes the cameras cover farther areas, hence this box is much larger than the area expected to be reconstructed. Therefore, another smaller box, which is determined by camera latitude and longitude range, is adopted to cover the central area, which is expected to be carefully reconstructed.

DistGrid uses a two-level cascade structure like NeRF++ \cite{zhang2020nerf++} but without inverse distance. The fine-level NeRF sets the inner box as its bounding box while the coarse-level NeRF uses the outer one. Region partitioning is performed based on the inner box, and the outer box is partitioned as well with the same boundaries. The hash table length of fine-level MHG scales up to $2^{24}$ for high-quality reconstruction, while the length of the coarse-level hash table is fixed to $2^{19}$ to ensure computing efficiency. An example of partitioning is illustrated in Fig.~\ref{fig:partition} under the condition of 1, 2, and 4 regions, respectively.

\definecolor{myred}{RGB}{178, 34, 34}
\definecolor{myblue}{RGB}{0, 191, 255}
\begin{figure}
    \centering
    \begin{subfigure}[b]{0.15\textwidth}
        \centering
        \includegraphics[width=\textwidth]{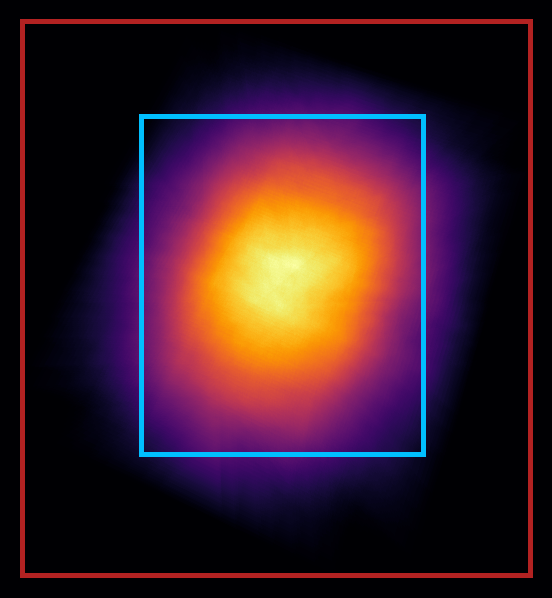}
        \caption{1 partition}
        \label{fig:partition1}
    \end{subfigure}
    \begin{subfigure}[b]{0.15\textwidth}
        \centering
        \includegraphics[width=\textwidth]{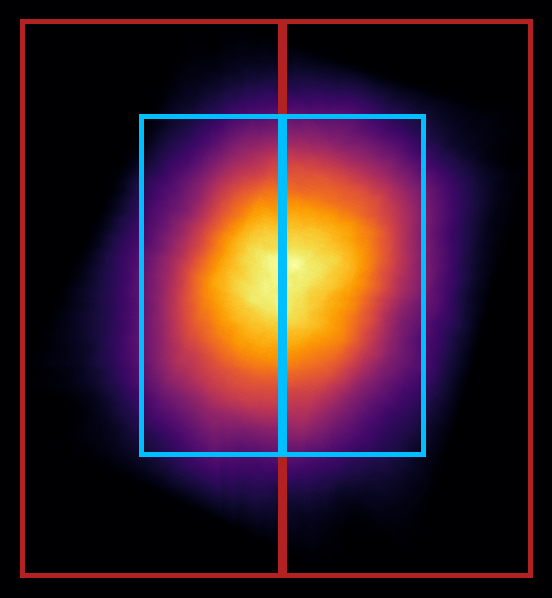}
        \caption{2 partitions}
        \label{fig:partition2}
    \end{subfigure}
    \begin{subfigure}[b]{0.15\textwidth}
        \centering
        \includegraphics[width=\textwidth]{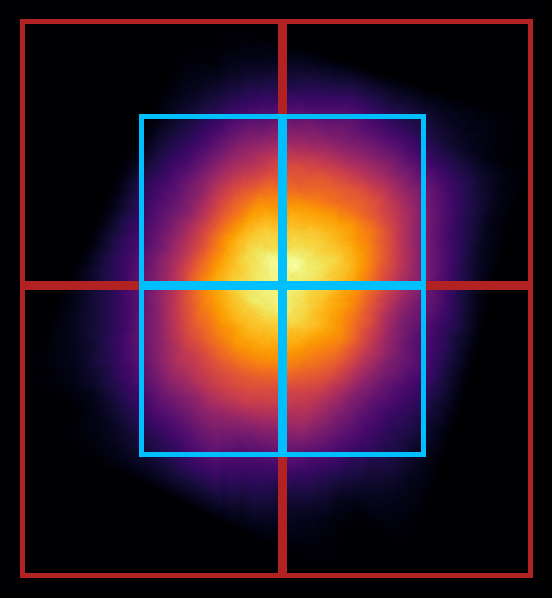}
        \caption{4 partitions}
        \label{fig:partition4}
    \end{subfigure}
    \caption{Two partitioning types described in sec.~\ref{sec:Partition}. All cameras' FOVs are projected onto the ground plane. Brighter regions are covered by more cameras. In coarse-fine partitioning, \textcolor{myblue}{fine-level box} wraps the region covered by most cameras, while \textcolor{myred}{coarse-level box} wraps the region covered by any of these cameras. With region partitioning, original scene~(\ref{fig:partition1}) can be partitioned into 2 regions~(\ref{fig:partition2}) or 4 regions~(\ref{fig:partition4}). }
    \label{fig:partition}
\end{figure}

\subsection{Segmented Volume Rendering}
\label{sec:SegmentedVren}

When performing volume rendering, the rays located in a single region can be rendered directly. However, the cross-region rays should be handled in a different way. To describe the proposed segmented volume rendering method, we extend the standard volume rendering (Eq.~\ref{eq:standard_vren}) with the parameterized start and end (Eq.~\ref{eq:vren}).

\begin{equation}
\label{eq:vren}
    \begin{aligned}
        C(\mathbf{r}) = C(\mathbf{r}|t_n, t_f) = \int_{t_n}^{t_f}T(t_n, t)\sigma(\mathbf{r}(t))c(\textbf{r}(t), \mathbf{d})dt, \\\text{where } T(t_n, t)=\exp\left(-\int_{t_n}^t(\sigma(\mathbf{r}(s))ds\right)
    \end{aligned}
\end{equation}

Now, assuming a ray $\mathbf{r}$ is crossing $K$ regions, entering $1^{st}$ region at time $t_0$, crossing the boundary between $i^{th}$ and $i+1^{th}$ region at time $t_i$, and leaving $K^{th}$ region at time $t_K$. We first utilize the integration-by-part to break the integration of transmittance into two parts. Given $t_j:t_i<t_j<t$, the transmittance $T(t_i, t)$ can be rewritten as Eq.~\ref{eq:transmittance}.

\begin{equation}
\label{eq:transmittance}
\begin{aligned}
    T(t_i, t) =& \exp\left(-\int_{t_i}^t(\sigma(\mathbf{r}(s))ds\right) \\
    =& \exp\left(-\int_{t_i}^{t_j}(\sigma(\mathbf{r}(s))ds-\int_{t_j}^t(\sigma(\mathbf{r}(s))ds\right) \\
    =& T(t_i, t_j)T(t_j,t)
\end{aligned}
\end{equation}

Next, we split the integration of Eq.~\ref{eq:vren} by part, and apply the results of Eq.~\ref{eq:transmittance}, then the Eq.~\ref{eq:color_fwd} is obtained.

\begin{equation}
\label{eq:color_fwd}
\begin{aligned}
    C(\mathbf{r}) =& C(\mathbf{r}|t_0, t_K) \\
    =& \int_{t_0}^{t_K}T(t_0, t)\sigma(\mathbf{r}(t))c(\textbf{r}(t), \mathbf{d})dt \\
    =& \sum_{i=0}^{K-1} \int_{t_i}^{t_{i+1}}T(t_0, t)\sigma(\mathbf{r}(t))c(\textbf{r}(t), \mathbf{d})dt \\
    =& \sum_{i=0}^{K-1} \prod_{j=-1}^{i-1}T(t_j,t_{j+1}) \int_{t_i}^{t_{i+1}}T(t_i, t)\sigma(\mathbf{r}(t))c(\textbf{r}(t), \mathbf{d})dt \\
    =& \sum_{i=0}^{K-1} \prod_{j=-1}^{i-1}T(t_j,t_{j+1}) C(\mathbf{r}|t_i,t_{i+1})
\end{aligned}
\end{equation}
where we define $T(t_{-1}, t_0) = 1$ for consistency. The partial color $C(\mathbf{r}|t_i,t_{i+1})$ and partial transmittance $T(t_i,t_{i+1})$ can be locally computed in $i+1^{th}$ region. The rendered color is composed with the Eq.~\ref{eq:color_fwd} and the final transmittance is computed through Eq.~\ref{eq:transmittance_fwd}. 

\begin{equation}
\label{eq:transmittance_fwd}
    T(t_0, t_K)=\prod_{i=0}^{K-1}T(t_i,t_{i+1})
\end{equation}

The color and transmittance are calculated in all $K$ regions and back-propagated only w.r.t the partial color and transmittance in each region. As the combination operator is not natively supported in mainstream Deep Learning frameworks, e.g., PyTorch~\footnote{https://www.pytorch.org}, the backward pass should be manually implemented. The backward pass of Eq.~\ref{eq:color_fwd} is described in Eq.~\ref{eq:color_bwd}.

\begin{equation}
\label{eq:color_bwd}
    \frac{\partial \mathcal{L}}{\partial C(\mathbf{r}|t_i,t_{i+1})} = \frac{\partial \mathcal{L}}{\partial C(\mathbf{r})}\frac{\partial C(\mathbf{r})}{\partial C(\mathbf{r}|t_i,t_{i+1})} = \frac{\partial \mathcal{L}}{\partial C(\mathbf{r})}\prod_{j=-1}^{i-1}T(t_j,t_{j+1})
\end{equation}

The training of DistGrid involves the loss about ray transmittance. Therefore the backward pass of Eq.~\ref{eq:transmittance_fwd} is required as well. The local transmittance is used for the composition of both $C(\mathbf{r})$ and $T(t_0, t_K)$ in the forward pass, so its gradient comes from these two parts, as is shown in Eq.~\ref{eq:transmittance_bwd}.

\begin{equation}
\label{eq:transmittance_bwd}
\begin{aligned}
    \frac{\partial \mathcal{L}}{\partial T(t_i,t_{i+1})} =& \frac{\partial \mathcal{L}}{\partial T(t_0, t_K)}\frac{\partial T(t_0, t_K)}{\partial T(t_i,t_{i+1})} + \frac{\partial \mathcal{L}}{\partial C(\mathbf{r})}\frac{\partial C(\mathbf{r})}{\partial T(t_i,t_{i+1})} \\
    = &\frac{\partial \mathcal{L}}{\partial T(t_0, t_K)}\prod_{j=-1 (j\neq i)}^{K-1}T(t_j,t_{j+1}) \\ 
    &+ \frac{\partial \mathcal{L}}{\partial C(\mathbf{r})}\sum_{k=i+1}^{K-1}\left[\prod_{j=-1 (j\neq i)}^{k-1}T(t_j,t_{j+1})\right]C(\mathbf{r}|t_k,t_{k+1})
\end{aligned}
\end{equation}

The segmented volume rendering is derived from the original volume rendering equation, which ensures that it can achieve consistent results from a single model in a distributed setting, without worrying about boundary artifacts. The integration-by-part is similar to the idea of foreground-background co-integration in NeRF++~\cite{zhang2020nerf++}, but the difference is that DistGrid gets rid of the background module and extends it to more than 2 segments.

\subsection{Training}
\label{sec:training}

\subsubsection{Training Data}

The input rays are uniformly sampled from all training images. Unlike Mega-NeRF which takes several hours to assign rays to corresponding chunks before training, our DistGrid requires no dataset preprocessing. All sub-NeRFs share the same data distribution. As the number of rays is over 20 billion~(e.g. about 1500 images with the resolution of 4608x3456 in dataset \textit{Rubble}), they can't all be loaded into memory. And sampling a batch from disk is heavily time-consuming. For this reason, we make a workaround that DistGrid maintains a ray cache in memory. The main process samples batch from this cache, while several concurrent background processes update the cache by loading images from the disk and sampling new rays to replace the old rays. The ray cache is distributed to all the sub-NeRF nodes, each of which loads and samples a portion of training images. 

\subsubsection{Training Procedure}

When a DistGrid sub-NeRF receives a batch of rays, it performs Ray-Intersection with AABBs to determine which regions these rays intersect, as well as the intersecting order, then sends these rays to their corresponding sub-NeRFs. After that, it receives a batch of rays that have an intersection with its bounding box. The ray marching and volume rendering is performed locally to obtain partial colors and transmittances. 
Once local volume rendering is complete, the partial colors and transmittances are scattered to other sub-NeRFs. Now the sub-NeRF has all the necessary variables to perform segmented volume rendering. After rendering, the ray colors and terminating transmittances are supervised with ground truth, and the backward pass is performed based on Eq.~\ref{eq:color_bwd} and Eq.~\ref{eq:transmittance_bwd}. The detailed training procedure for a single cross-region ray is illustrated in Fig~\ref{fig:seg_vren}.

Evaluation is performed in a master-slave mode, where the master sub-NeRF dispatches rays to corresponding slave sub-NeRFs, and collects their partial rendering results. The results are then merged with segmented volume rendering to obtain the final color, depth, and transmittance.

\subsubsection{Appearance Features}

The appearance features are used in the color network to provide DistGrid with additional flexibility in explaining lighting differences across images. Unlike trainable appearance embeddings in NeRF-W~\cite{martin2021nerf}, the appearance features for each image are fixed. The appearance features come from VGG's~\cite{simonyan2014very} first layer feature maps. First $c$ channel features are extracted to compute its gram matrix~\cite{mordvintsev2015deepdream} of shape $c\times c$. Then the dimension of these gram matrices is reduced to appearance feature dimensions using Principal Component Analysis (PCA). The PCA algorithm requires that $c^2$ is smaller than the dataset size. The fixed appearance features make it easier to evaluate the quality of novel views from the validation set.

The images captured by drones can be foggy or light-inconsistent. Volume rendering has two ways to account for these artifacts in the scene. The easier way is to treat them as floaters, which causes ghostly artifacts, while the harder way is to interpret them as a filter of the image with the help of appearance features, which are expected. Therefore, the model is regularized to reduce floaters through an aggressive pruning strategy. Specifically, a relatively large density threshold is applied to Instant NGP's occupancy grid \citep[app. E.2]{muller2022instant} to prune semitransparent areas, so that the appearance features are activated to make the rendered view light-consistent. 

\subsubsection{Loss Functions}

We adopt the mean square error~(MSE) between the rendered color $C(\mathbf{r})$ and the ground-truth color $C_{gt}(\mathbf{r})$ as the loss function:
\begin{equation}
\label{eq:l_rgb}
    \mathcal{L}_{rgb} = \sum_\mathbf{r} ||C(\mathbf{r})-C_{gt}(\mathbf{r})||^2_2.
\end{equation}

In the oblique photography scenario, all the rays should terminate within the bounding box, so the final transmittance of the rays should get close to zero. Thus, a transmittance regularizer is added to the loss function:
\begin{equation}
\label{eq:l_transmittance}
    \mathcal{L}_{T} = -\sum_\mathbf{r} \log\left(1-T(t_n, t_f|\mathbf{r})\right).
\end{equation}

The distortion loss $\mathcal{L}_{dist}$, which is firstly proposed in MipNeRF-360~\cite{barron2022mip}, is further utilized to encourage volume rendering weights to be compact and sparse, alleviating floater or background collapse artifact. Loss weights $\lambda_1$ and $\lambda_2$ are used to scale later two losses, and the overall loss function is:
\begin{equation}
\label{eq:loss}
\mathcal{L} = \mathcal{L}_{rgb} + \lambda_1 \mathcal{L}_{T} + \lambda_2 \mathcal{L}_{dist}.
\end{equation}

We set $\lambda_1=\lambda_2=10^{-3}$ in all our experiments.

\section{Experiments}

\def\psnr{$\uparrow$PSNR}
\def\ssim{$\uparrow$SSIM}
\def\lpips{$\downarrow$LPIPS}
\begin{table*}[t]
\caption{The comparison to NeRF, TensoRF, Mega-NeRF, GP-NeRF, and Grid-Guided NeRF on the metrics of PSNR, SSIM, and LPIPS. The configuration of $2^{24}\times 4$ is used in DistGrid. DistGrid consistently outperforms the baselines. $^*$LPIPS reported in Grid-Guided NeRF is not implemented with VGG.}
\resizebox{0.8\textwidth}{!}{%
\begin{tabular}{c|ccc|ccc|ccc|ccc}
\toprule
Scene & \multicolumn{3}{c|}{\textit{Rubble}} & \multicolumn{3}{c|}{\textit{Building}} & \multicolumn{3}{c|}{\textit{Campus}} & \multicolumn{3}{c}{\textit{Residence}} \\ \midrule
Metric                             & \psnr  & \ssim & \lpips& \psnr  & \ssim & \lpips& \psnr  & \ssim & \lpips& \psnr  & \ssim & \lpips\\ \midrule
NeRF~\cite{mildenhall2021nerf}     & 21.428 & 0.424 & 0.662 & 18.795 & 0.392 & 0.663 & 20.707 & 0.426 & 0.702 & 18.147 & 0.419 & 0.682 \\
TensoRF~\cite{chen2022tensorf}     & 22.835 & 0.587 & 0.478 & 20.443 & 0.567 & 0.466 & 21.238 & 0.553 & 0.570 & 19.812 & 0.621 & 0.454 \\
Mega-NeRF~\cite{turki2022mega}     & 24.060 & 0.553 & 0.516 & 20.930 & 0.547 & 0.504 & 23.420 & 0.537 & 0.618 & 22.080 & 0.628 & 0.489 \\
GP-NeRF~\cite{zhang2023efficient}  & 24.080 & 0.563 & 0.497 & 20.990 & 0.565 & 0.490 & 23.460 & 0.544 & 0.611 & 22.410 & 0.659 & 0.451 \\
Grid-Guided NeRF~\cite{xu2023grid} & 25.467 & 0.780 & 0.213$^*$ & -      & -     & -     & 25.505 & \textbf{0.767} & 0.174$^*$ & 24.372 & 0.807 & 0.142$^*$ \\ \midrule
DistGrid (Ours)                     & \textbf{28.189} & \textbf{0.819} & \textbf{0.233} & \textbf{24.434} & \textbf{0.760} & \textbf{0.302} & \textbf{26.485} & 0.737 & \textbf{0.319} & \textbf{25.273} & \textbf{0.810} & \textbf{0.277} \\ \bottomrule
\end{tabular}%
}
\label{tab:evaluation}
\end{table*}

\subsection{Experiment Setup}

\textbf{Dataset.} Following Mega-NeRF, we use 4 real-world urban scenes in our experiments, named \textit{Rubble}, \textit{Building}, \textit{Campus} and \textit{Residence}~\cite{turki2022mega,lin2022capturing} respectively. The camera poses are estimated with Pixel-Perfect SfM~\cite{lindenberger2021pixel}.

\noindent\textbf{Implementations.} The DistGrid is implemented based on a Python-implemented Instant NGP\footnote{https://github.com/kwea123/ngp\_pl}. The hash-grid resolution is fixed to 8192, as it's found to be irrelevant to performance. Adam~\cite{kingma2014adam} optimizer is used with a learning rate initialized to 0.05 and scaled down to 0.005 with a cosine scheduler during 300$k$ training iterations. The appearance feature dimension is set to 16 for \textit{Rubble} and \textit{Building}, 48 for \textit{Campus} and \textit{Residence}. To stabilize the training of coarse-level MHG, $sigmoid$ is used instead of $ReLU$ as the activation function in color MLP, and the outputs of both the density and color MLPs are clipped within $[-15,15]$ before output activations. 

The occupancy-grid-guided sampling is used instead of hierarchical sampling, thus an occupancy grid is maintained during training. Unlike~\cite{muller2022instant}, the multi-scale setting and Morton storage order are disabled for compatibility. The occupancy grid has a resolution of 512, and the density value decays by a factor of 0.99 after every 16 training iterations. The warm-up stage is extended to 4096 steps. The density threshold for occupancy bit-field is 0.6 at the first 10$k$ iterations and is 60 after that. The hyper-parameters not mentioned here remain the same as~\cite{muller2022instant}. 

The in-memory ray cache has a size of $10^8$, and 4 processes in each sub-NeRF update the ray cache in the background, sampling $10^8\times K/N$ rays for each image, where $N$ is the dataset size and $K$ is the partition count. The batch size is set to 8192. All experiments are carried out on a device with 4 PCIe-connected RTX 3090~(24G) GPUs. 

\noindent\textbf{Baselines.} Our DistGrid is compared with basic NeRF~\cite{mildenhall2021nerf}, TensoRF~\cite{chen2022tensorf}, Mega-NeRF~\cite{turki2022mega}, GP-NeRF~\cite{zhang2023efficient}, and Grid-Guided NeRF~\cite{xu2023grid}. For NeRF implementation, the neural field is represented by an MLP with 12 layers and 256 hidden units. The frequency of position encoding varies from $2^0$ to $2^{15}$, inserted to MLP via skip connection at the 4th and 8th layers. 64 coarse and 128 fine samples are sampled per ray using hierarchical sampling. The learning rate is $5\times 10^{-4}$ and the batch size is 2048. The model is trained for 300$k$ steps. For TensoRF implementation, we use VM-192 and keep most of the hyper-parameters unchanged but increase the final voxel count to $1024^3$. Model is trained for 300$k$ steps and the grid is upsampled at step $[10k, 15k, 20k, 30k, 40k]$. For Mega-NeRF, GP-NeRF, and Grid-Guided NeRF, we directly compare the metrics reported in their papers, as the training of Mega-NeRF and GP-NeRF costs over 1TB of storage space per scene, which is not available for our device, while Grid-Guided NeRF hasn't released their source code. 

\noindent\textbf{Evaluation Metrics.} We evaluate the methods above in terms of PSNR, SSIM~\cite{wang2004image}, and the VGG implementation of LPIPS~\cite{zhang2018unreasonable}.


\subsection{Evaluation}

The performance of our DistGrid is compared to the baselines qualitatively and quantitatively. The DistGrid with $T=2^{24}$ and 4 partitions are used in the quantitative evaluation. As shown in Tab.~\ref{tab:evaluation}, DistGrid outperforms the baselines on all 4 large-scale datasets.

\begin{figure*}
    \centering
    \includegraphics[width=0.92\textwidth]{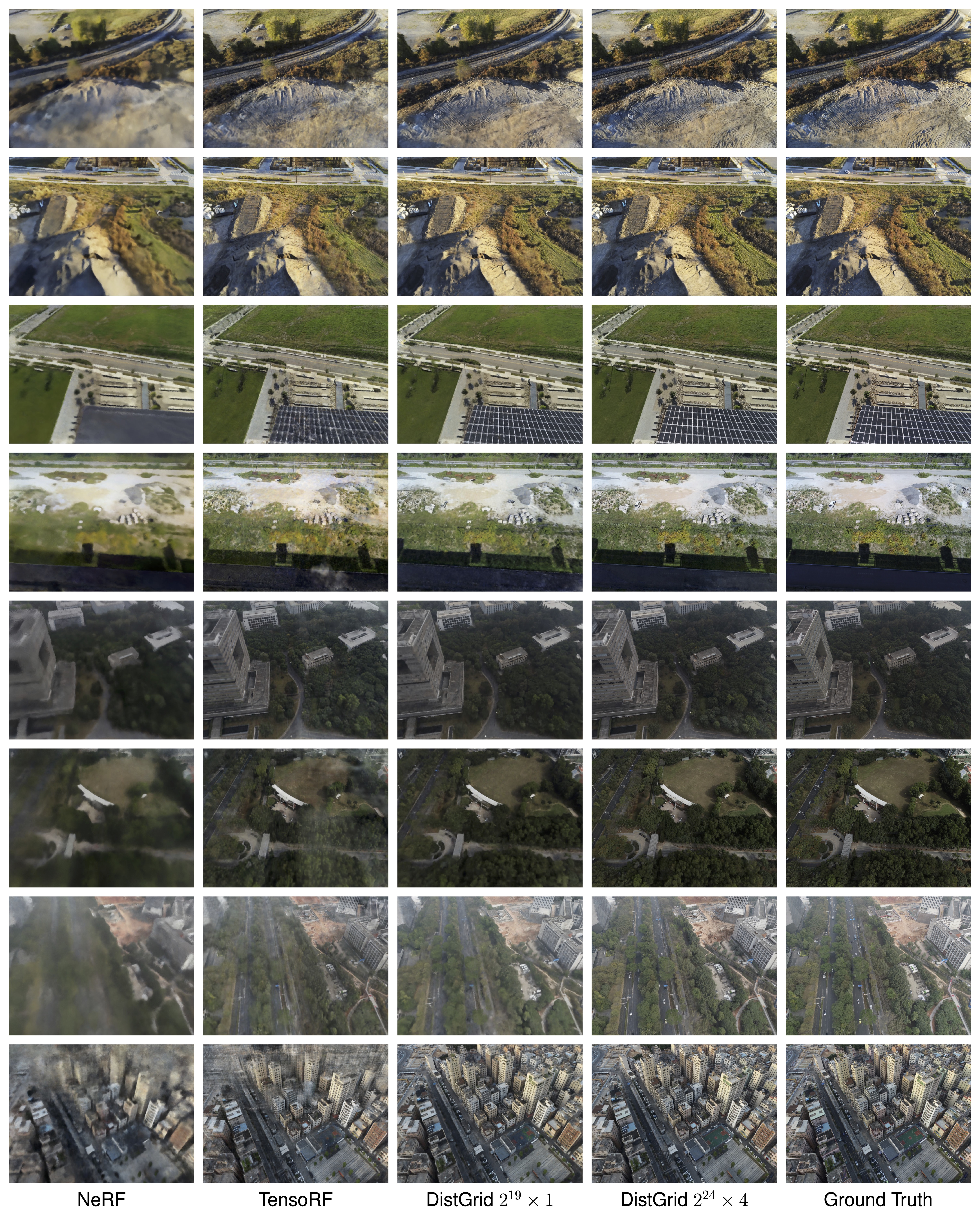}
    \caption{Qualitative evaluation on DistGrid, compared to NeRF and TensoRF.}
    \label{fig:evaluation}
\end{figure*}

\begin{figure*}
    \centering
    \includegraphics[width=0.96\textwidth]{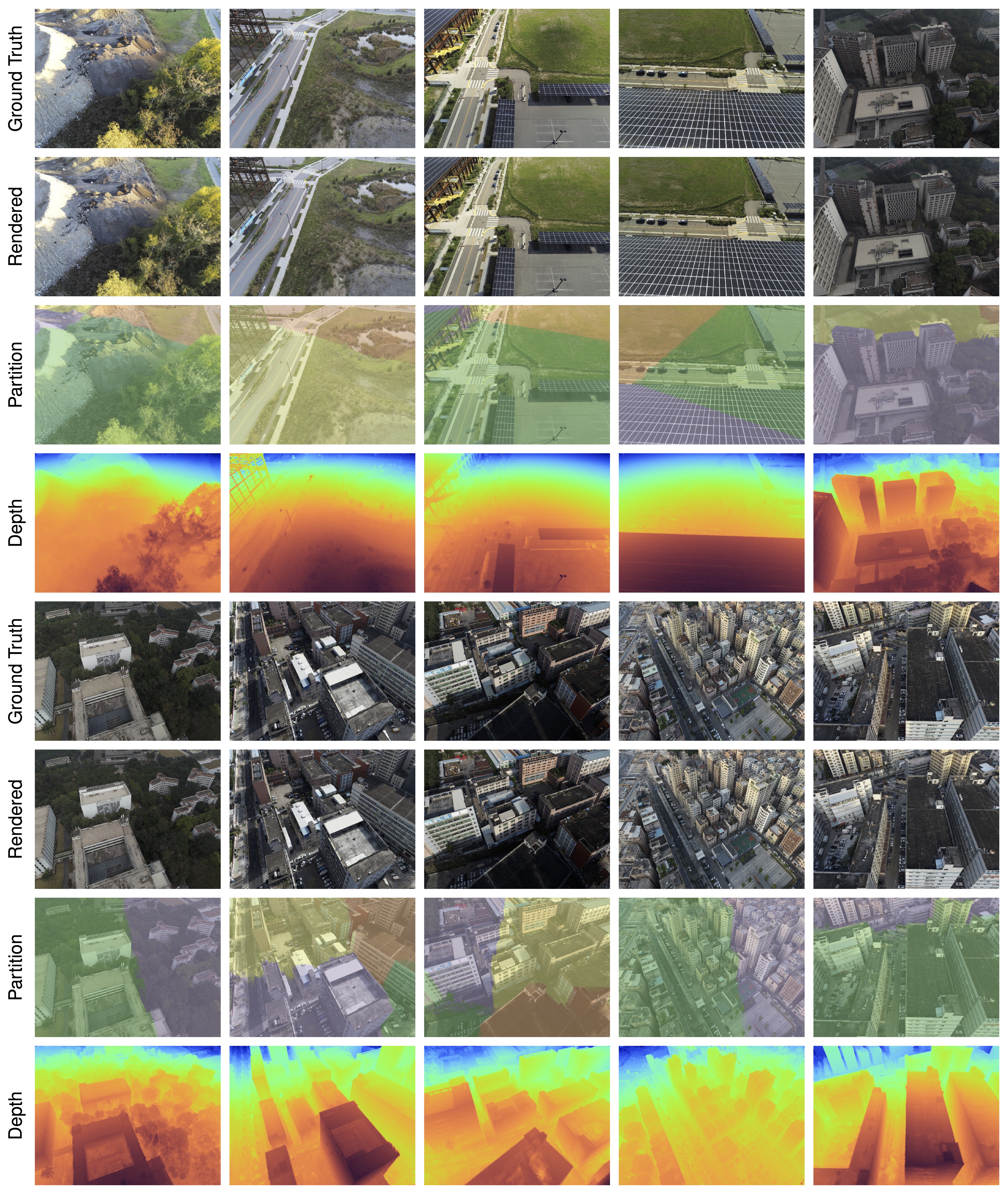}
    \caption{The rendered views on the boundary.}
    \label{fig:boundary}
\end{figure*}

Some rendered novel views are depicted in Fig.~\ref{fig:evaluation}. Compared to MLP-based NeRF~\cite{mildenhall2021nerf} and multi-planar-based TensoRF~\cite{chen2022tensorf}, our method achieved more realistic visual effects and rendered more high-frequency details. Scene partitioning in DistGrid is non-overlapped, thus we further evaluate the visual quality on the boundary of adjacent partitions in both the RGB domain and depth domain, as shown in Fig.~\ref{fig:boundary}. In the $3^{rd}$ row, pixels with different colors are rendered by different sub-NeRFs. In specific, each partition is assigned a unique color, and the pixel color is the weighted sum of these colors, where the weight is defined as the loss of transmittance after the ray passes through the region. With the help of the proposed segmented volume rendering, there is no perceived artifact at the boundary of adjacent regions. The depth variation at the boundary is also continuous.


\begin{figure}
    \centering
    \begin{subfigure}[b]{0.23\textwidth}
        \centering
        \includegraphics[width=\textwidth]{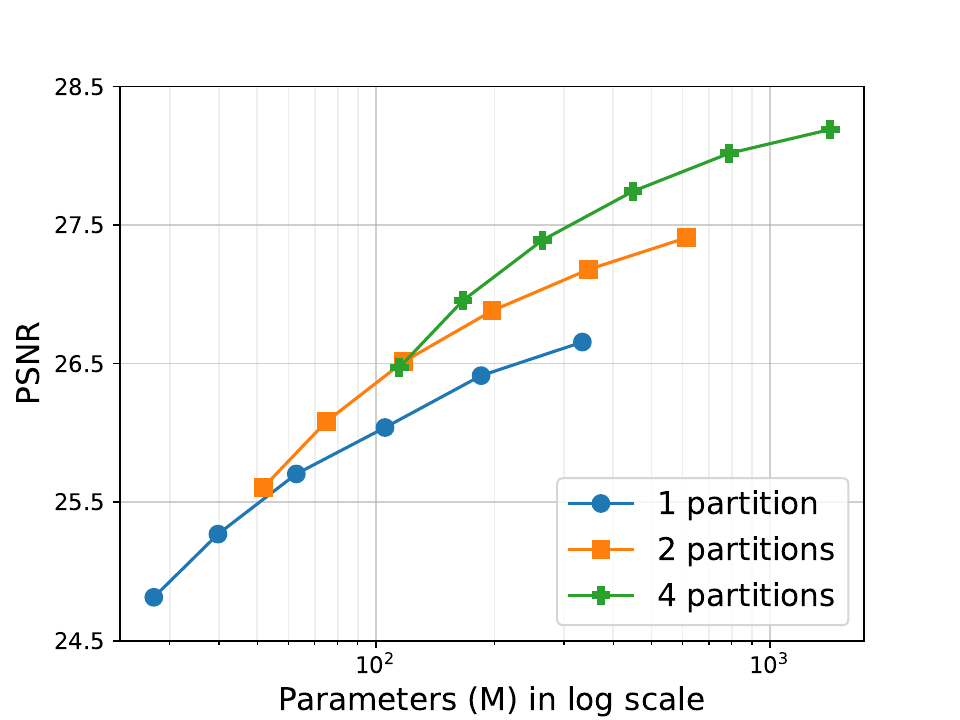}
        \caption{Parameter count to PSNR}
        \label{fig:rubble_psnr}
    \end{subfigure}
    \begin{subfigure}[b]{0.23\textwidth}
        \centering
        \includegraphics[width=\textwidth]{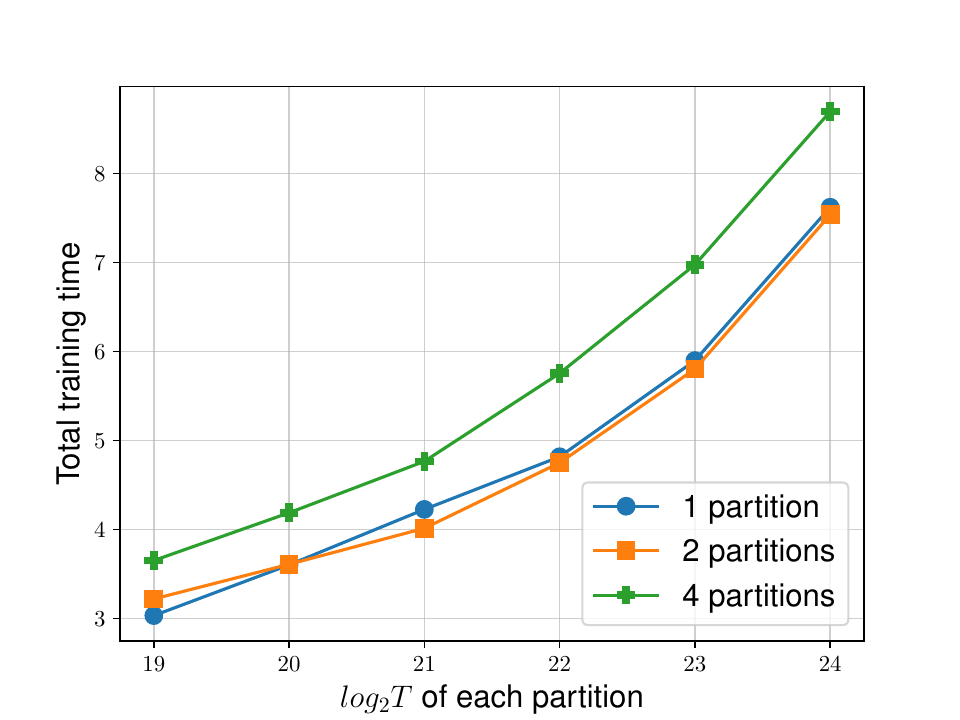}
        \caption{Hash table length to training time}
        \label{fig:rubble_time}
    \end{subfigure}
    \caption{Quantitative results on \textit{Rubble} with 1, 2, and 4 partitions. The hash table length ranged from $2^{19}$ to $2^{24}$.}
    \label{fig:rubble_results}
\end{figure}

\subsection{Scalability}

We then evaluate the scalability of the proposed DistGrid method. As shown in Fig.~\ref{fig:rubble_plot}, the performance improves with a larger hash table. Even though the hash tables are distributed across different GPUs, the growth trend is maintained. In this section, we analyze the effect of the number of parameters and the number of partitions on model performance and training time. Scene \textit{Rubble} is used for evaluation in this section. The scene is partitioned into 1, 2, and 4 regions like Fig.~\ref{fig:partition}. For each partition strategy, the hash table length varies from $2^{19}$ to $2^{24}$. The results are plotted in Fig.~\ref{fig:rubble_results}.

As shown in Fig.~\ref{fig:rubble_psnr}, in all partition strategies, the PSNR improves with the number of parameters, showing the scalability of the DistGrid method. An interesting finding is that with the same amount of, even fewer parameters, increasing partitions have a positive effect on model performance. For example, the model with $T=2^{24}\times1$ has 334.1M parameters and its PSNR achieves 26.7 dB; while the model with $T=2^{21}\times4$ has 264.6M parameters but its PSNR achieves 27.4 dB. 

As shown in Fig.~\ref{fig:rubble_time}, the ray exchange overhead can be ignored when scaled to 2 partitions. The training time increases when scaled to 4 partitions, but the largest model can still be trained in 9 hours.

\subsection{Ablation Study}

\begin{table}[t]
\caption{Ablation study on coarse-fine partitioning on dataset \textit{Rubble}. Average PSNRs are reported.}
\label{tab:coase-fine partitioning}
\resizebox{0.4\textwidth}{!}{%
\begin{tabular}{c|ccc}
\toprule
Hash table length & $2^{24}\times1$   & $2^{24}\times2$   & $2^{24}\times4$ \\ \midrule
\textit{w/o} Coarse-Fine Partitioning & 25.804 & 26.325 & 27.580 \\
\textit{w/} Coarse-Fine Partitioning  & 26.655 & 27.409 & 28.189 \\ \bottomrule
\end{tabular}%
}
\end{table}

In this section, we perform an ablation study on coarse-fine partitioning to prove its effectiveness. For comparison, the two-level bounding boxes are replaced with the outer coarse-level box. The comparison results on 3 scene partition types are shown in Tab.~\ref{tab:coase-fine partitioning}. The coarse-fine partitioning helps the fine-level model focus more on critical central areas, therefore performs better on scene reconstruction. As all the cameras are within the fine-level bounding box, the reconstruction of the outer region is more like an easier front-facing task. Besides, the outer scene is far from the cameras. For these reasons, the reconstruction of such regions can be handled by a coarse-level model with a relatively small hash table length.

\section{Conclusion and Discussion}

This work focuses on large-scale scene reconstruction. We propose a scalable scene reconstruction method based on joint MHGs, named DistGrid. This approach overcomes the memory limitation of previous volume-based large-scale scene reconstruction methods and eliminates the training redundancy of previous scene partition strategies. Our method shows high visual fidelity and scalability. 

While this work adopts MHG to represent sub-regions, it also suits any other methods whose bounding box is an AABB. Our DistGrid is currently trained on oblique photography datasets collected by drones. But it has the potential to reconstruct from multiple types of data sources, such as cellphone video, LiDAR sensors, in-vehicle cameras, or even remote sensing images. For such a large and hybrid reconstruction scope, our DistGrid needs to be scaled to a larger level, and a more efficient training pipeline should be designed, which will be discussed in the future.

\bibliographystyle{ACM-Reference-Format}
\bibliography{sample-bibliography.bib}

\end{document}